# Hybrid Approach for Electricity Price Forecasting using AlexNet and LSTM

**Bosubabu Sambana**
Assistant Professor,
Department of Computer Science and Engineering (Data Science),
School of Computing,
Mohan Babu University, Tirupati, India.
bosubabu.s@mbu.asia

**Kotamsetty Geethika Devi**
UG Scholar, Department of Computer Science and Systems Engineering,
Sree Vidyanikethan Engineering College,
Tirupati, India.
geethikakotamsetty@gmail.com

**Bandi Rajeswara Reddy**
UG Scholar, Department of Computer Science and Systems Engineering,
Sree Vidyanikethan Engineering College,
Tirupati, India.
rajeswarareddy03@gmail.com

**Galeti Mohammad Hussain**
UG Scholar, Department of Computer Science and Systems Engineering,
Sree Vidyanikethan Engineering College,
Tirupati, India.
hussaingaleti@gmail.com

**Gownivalla Siddartha**
UG Scholar, Department of Computer Science and Systems Engineering,
Sree Vidyanikethan Engineering College,
Tirupati, India.
gownivallasiddartha143@gmail.com

**Abstract:** The recent development of advanced machine learning methods for hybrid models has greatly addressed the need for the correct prediction of electrical prices. This method combines AlexNet and LSTM algorithms, which are used to introduce a new model with higher accuracy in price forecasting. Despite RNN and ANN being effective, they often fail to deal with forex time sequence data. The traditional methods do not accurately forecast the prices. These traditional methods only focus on demand and price which leads to insufficient analysis of data. To address this issue, using the hybrid approach, which focuses on external variables that also effect the predicted prices. Nevertheless, due to AlexNet's excellent feature extraction and LSTM's learning sequential patterns, the prediction accuracy is vastly increased. The model is built on the past data, which has been supplied with the most significant elements like demand, temperature, sunlight, and rain. For example, the model applies methods, such as minimum-maximum scaling and a time window, to predict the electricity prices of the future. The results show that this hybrid model is good than the standalone ones in terms of accuracy. Although we got our accuracy rating of 97.08, it shows higher accompaniments than remaining models RNN and ANN with accuracies of 96.64 and 96.63 respectively.

**Keywords**: Electricity Price Forecasting, Hybrid Model, AlexNet, LSTM (Long Short-Term Memory), RNN (Recurrent Neural Network), ANN (Artificial Neural Network), Time-Series Prediction

1. ## INTRODUCTION

Electricity price forecasting is extremely vital for energy markets, because precise forecasting is advantageous for electricity producers, traders, and consumers. It helps in risk management, operational improvement, and policy decision-making. On the other hand, the traditional forecasting methods are not as reliable as they should be due to the high volatility and the unpredictable events that are created by the variations in the weather, the prices of commodities, and the supply and demand of the market. Even though they have been commonly used, the models such as ANNs and RNNs fail to identify long-term dependencies and to mine useful features from huge datasets. To take long term dependencies we use LSTM. Such hybrid approaches that merge multiple machine learning methods have also been proposed as a solution to the above challenges. Especially, the inclusion LSTM for time-series forecasting along with the application of the AlexNet model for obtaining features is seen as a promising solution to the shortcomings of individual models. In this research paper, we introduce a hybrid theory by combining AlexNet with LSTM. Although LSTM captures long-term patterns in electric charges, AlexNet, on the other hand, is used to extract relevant features from the input data. A dataset from Kaggle, which has various elements such as demand, temperature, and sunlight intensity, is the one used to train this model. To make the data more adjustable, Min-Max scaling is applied and the sliding window technique is ideal to account for the premise of time series and allow perfect prediction. Our hybrid solution is a more stable and flexible tool for price forecasting by giving a horizon of up to 72 months and also, we can predict for the upcoming years using the model with the LSTM algorithm. This idea comes in handy in fact because it not only takes care of accuracy in forecasting but also a feature that creates a personal future price projection system.

## 2. LITERATURE SURVEY

The more modern machine learning (ML) techniques and hybrid approaches were explored with in addition to traditional time series modelling approaches in order to increase forecast accuracy. Different methodologies were considered in the literature on electricity price prediction. These are helpful to address the issues in complex power markets.

Popeanga and Lungu et.al state that time series analysis and the centred moving average approach are used to forecast energy usage. Their work thus highlights the potential role that uncovering underlying patterns in past data may have in producing such trustworthy forecasts [1].

Ostertagová and Ostertag et.al have shown, through the use of exponential smoothing approaches, that basic smoothing techniques can be highly useful in short-term power price forecasting; nevertheless, one potential drawback of these techniques could be their failure to capture abrupt price spikes [2].

Nazim and Afthanorhan et.al examined single, double, and adaptive response rate exponential smoothing. Their research showed that adaptive methods—while more complicated—performed better in situations of fast change and, as a result, are suitable for use in markets for electricity, where price volatility is common [3].

Abd Jalil et al, reported an application of exponential smoothing techniques in electricity load demand forecasting, highlighting the applicability of such techniques for markets with mild volatility [4].

Pedregal and Trapero et.al employed a multi-rate technique to anticipate electricity in the mid-term, highlighting the need for dynamic models that can adapt to different time intervals [5].

Almeshaiei and Soltan et.al proposed an approach to electric power load forecasting. It dealt with integrating several forecasting techniques to adapt to new variations in demand loads brought on by weather and consumer behaviour [6].

Muhamad and Mohamed Din et.al apply exponential smoothing techniques to time series data on levels of the water in the river. Their work demonstrates the effectiveness of these methods in forecasting environmental data, particularly in hydrological studies, where accurate predictions are crucial for managing water resources [7].

Tirkeş, Güray, and Çelebi present a comparison of demand forecasting methods, including Holt-Winters, Trend Analysis, and Decomposition Models. Their findings highlight the strengths and weaknesses of each method in forecasting demand, offering valuable insights into their application in various industries, particularly for energy and utilities forecasting [8].

Kavanagh et.al explores short-term demand forecasting for the Integrated Electricity Market. The paper discusses forecasting models designed to predict electricity consumption, an essential task for ensuring the stability and efficiency of energy distribution in competitive markets [9].

The limitations of these existing models are they do not include the possible features that can affect the electricity price. They tend to ignore the external factors like solar exposure or rainfall and solely focus on the price and the electricity demand for the analysis. To address this, we included many features that effect the electricity price, also included the external factors, seasonal and weather patterns. The key challenge is data preprocessing and the feature engineering, this challenge was addressed using various methods like handling the missing values, replacing the missing values with the mean, median or previous values. also, feature engineering is performed by capturing the seasonal patterns, scaling and normalizing techniques.

Hybrid models that take advantage of several other forecasting strategies emerged as effective in accommodating the shortcomings of individual methods. Hybridization of the deep learning architecture such as AlexNet into LSTM for feature extraction to capture its time relationship will form these promising models. This type of model benefits both the strengths of convolutional neural networks and recurrent neural networks in producing better accuracy when there are volatile markets or high spiking electricity prices. The sophisticated techniques of the advanced machine learning that derived from traditional time-series are used for enhanced electricity price forecasting. By combining AlexNet and LSTM, hybrid models can overcome the shortcomings of current approaches, offering better accuracy in dynamic electricity markets. This improvement aids in decision-making and risk management for energy trading and consumption. To further improve forecasting performance in the energy sector, more research into hybrid models is necessary due to the increasing complexity of electricity price patterns.

### 3. METHODOLOGY

As shown in the Figure-1, the very first step consists of retrieving a dataset from Kaggle. The dataset is cleaned and pre-processed for the purpose of the model training to cater to data such as outliers, presence of null values, etc. The features are pre-processed using the min-max scaling to scale between 1 and 0. Also the missing values are handled by replacing with previous values because they are related to the time series data. The dataset is then divided into two sets-70% of data for training and 30% of data for testing purposes. The time series forecasting is performed by ALEXNET and LSTM, where AlexNet is mainly used for feature extraction from dense layers, and the second one is LSTM for analysing sequential data based on some temporal dependencies present in the input sequences. The features are picked up and trends are recognized with LSTM- how the past demand can dictate present pricing or something like that. Finally, the model performance is validated using RMSE and MAE performance measures. Once evaluated it outputs the predicted future electricity prices.

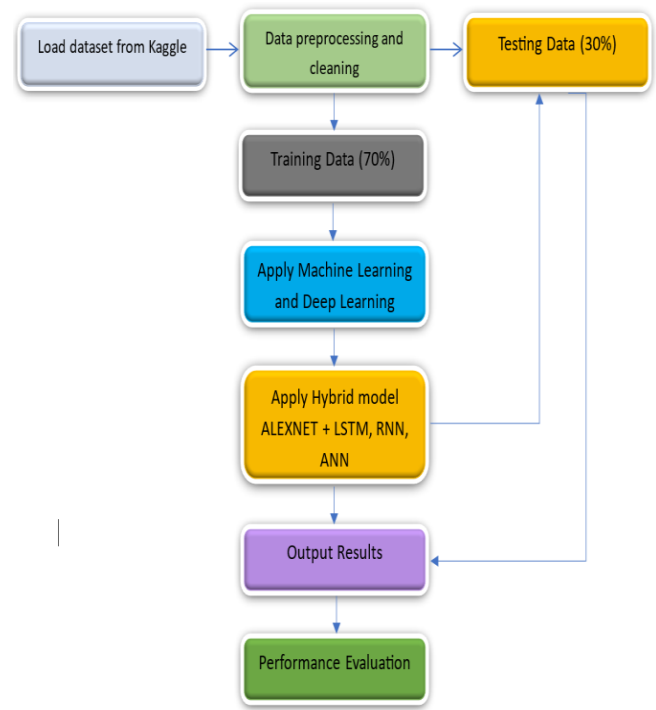

**Figure.1** Architecture of Electricity Price Prediction Model

### 4. IMPLEMENTATION

The project's dataset was taken from the Kaggle, and the dataset contains over 2000 entries for every feature in it. The dataset has Date, Demand, RRP(Price), Solar Exposure, Max Temperature, Min Temperature, RRP Positive, RRP Negative and etc. The size of the file is about 233 KB; hence it is proceeded for the analysis and further processing. Any Null values or undefined values are erased with the data cleaning and the data preprocessing process. After which, the dataset is divided into training data and testing data ,70% and 30% respectively. Now both the hybrid and the conventional methods are applied on the pre-processed. The conventional methods are ANN and RNN and our Hybrid model that is AlexNet-LSTM. After the application of methods, the models' performance is evaluated using the RMSE and MAE metrics. These metrics are used to check the performance of the model. Compare the overall performance of different conventional and machine learning algorithms used to identify the better approach.

$$RMSE = \sqrt{\frac{1}{n}\sum_{j=1}^{n}(y_j - \hat{y}_j)^2} \quad \ldots (1)$$

Where:
    n = The combined count of data values
    yj= actual value
    y^j = predicted value

$$MAE = \frac{1}{n}\sum_{j=1}^{n}|y_j - \hat{y}_j| \quad \ldots (2)$$

Where:
    n = The combined count of data values
- yj= actual value
- y^j = predicted value

## 5. ANALYSIS OF EXPERIMENTAL RESULTS

The model designed is the combination of AlexNet with the LSTM. This model was contrasted with more conventional models like ANN and RNN. The accuracy, recall, precision, and the F-score being the main metrics used in the testing process. See Table 1 where the AlexNet + LSTM hybrid model exceeded all other models by attaining the highest accuracy (97.08%), precision (0.30), recall (0.96), and F-score (0.43) (as shown in Table -1). However, the model being discussed was successful in capturing both time series and spatial features, making it a powerful prediction tool for electricity prices. The LSTM model, on the other hand, in its effort of grabbing independence with time variables, did reasonably well, but it did even better once integrate with AlexNet. To further enhance the performance, fine tuning the hyper parameters of LSTM model, adjusting the hidden layers, batch size for optimal results.

**Table. 1** Performance Comparison of the LSTM+ALEXNET, ANN and RNN algorithms

| Algorithm | LSTM+ ALEXNET | RNN | ANN |
|---|---|---|---|
| Accuracy (%) | 97.08 | 96.64 | 96.63 |
| Precision | 0.28 | 0.30 | 0.22 |
| Recall | 0.96 | 0.49 | 0.43 |
| F-Score | 0.43 | 0.37 | 0.29 |

**Accuracy:**

$$accuracy = \frac{(true\ positives + true\ negatives)}{Total\ no\ of\ Test\ Samples} \quad \ldots (3)$$

**Precision:**

$$Precision = \frac{(True\ Positives)}{True\ Positives + False\ Positives} \quad \ldots (4)$$

**Recall:**

$$Recall = \frac{(True\ Positives)}{True\ Positives + False\ Negatives} \quad \ldots (5)$$

**F-Score:**

$$F - Score = 2X\frac{(Precision * Recall)}{(Precision + Recall)} \quad \ldots (6)$$

Accuracy is defined as the expected overall correctness of predictions. This is obtained by calculating the ratio of correct predictions to any total predictions. To achieve high accuracy data preprocessing should be effective also by fine-tuning the model by adjusting the parameters like number of layers. This hybrid model gives good accuracy (as shown in Figure 2) because it follows external factors seasonality and weather trends.

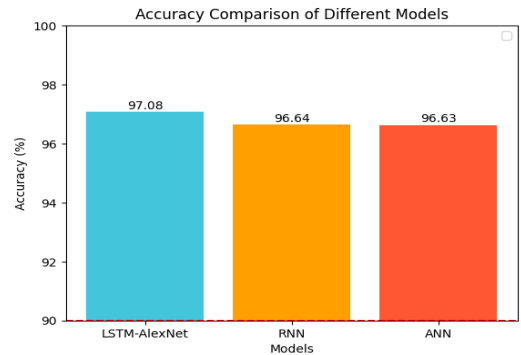

**Figure.2** Accuracy Comparison of Models

Precision measures how many positive predictions were correct. To enhance the precision, an optimal window size needs to be selected. Since the data here is dealing with time series, increasing the window size allows to capture long-term patterns. Too long window can cause additional noise so required span is important to enhance the precision. (From the below Figure -3) The precision of Hybrid model scored good compared to other models.

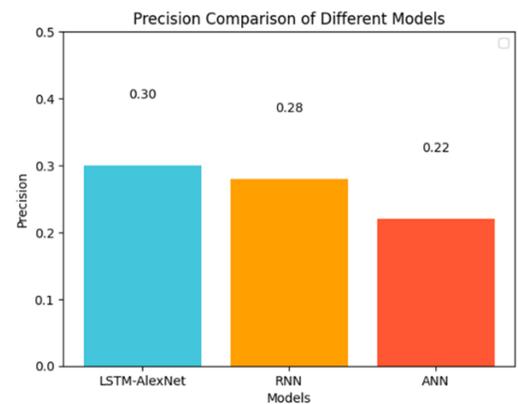

**Figure. 3** Precision Comparison of Models

Recall measures how well the classifier identifies all relevant instances, with a high recall indicating that not many are missed. F-Score combines precision and recall to arrive at a single metric to which one can appeal in balancing the trade-off. Relevant when both false positives and false negatives carry weight. (as shown in figure-4 and figure-5)

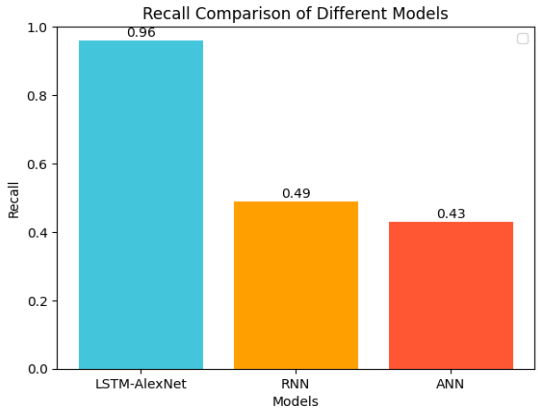

**Figure.4** Recall Comparison of Models

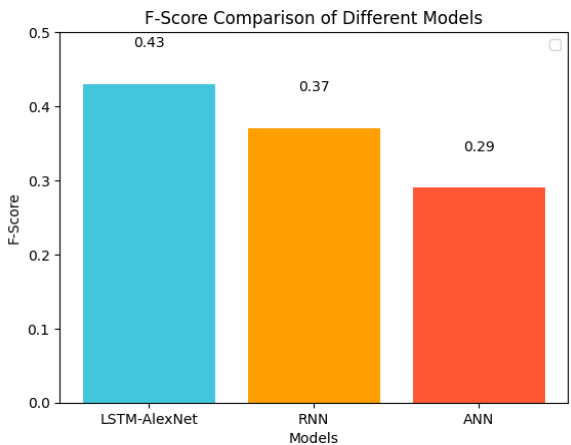

**Figure.5** F-Score Comparison of Models

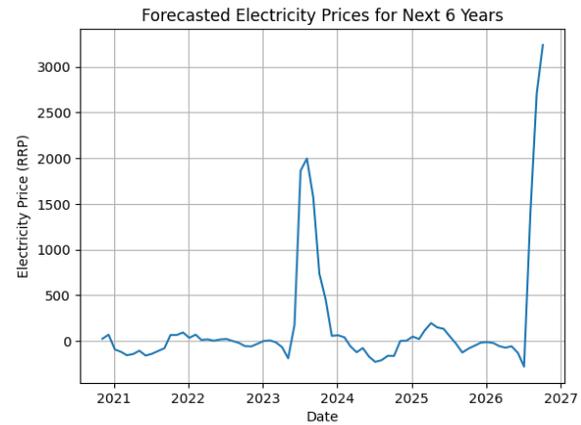

**Figure.6** Forecasted Electricity Prices for Next 6 Years

The graph shown in Figure -6 is a line plot forecasting electricity prices (measured as RRP, or Regional Reference Price) over a period from 2021 to 2027. The y-axis represents the electricity price, while the x-axis represents the years.

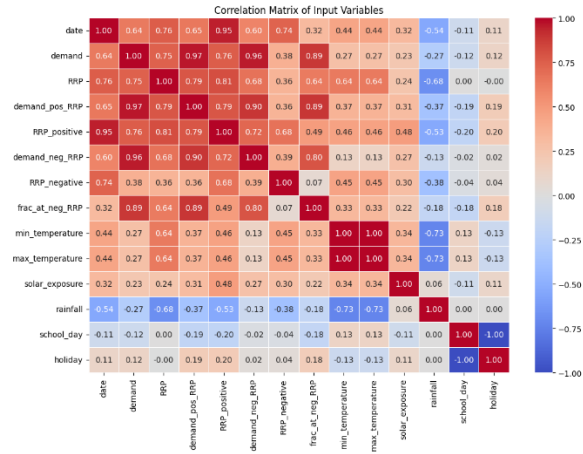

**Figure.7** Correlation matrix of input variables

(As shown in above figure 7) correlation analysis is performed to study the relationship between various input features. These features influence the forecasting. The hybrid model was able to identify strong correlation between primary inputs like demand and RRP. Also, correlation is shown between categorical features like holiday and external factors like solar exposure, rainfall etc.

## 6. CONCLUSION AND FUTURE WORK

This study presents a hybrid model combining AlexNet and LSTM for electricity price prediction, effectively capturing both long- and short-term patterns in price data. By leveraging AlexNet's spatial feature extraction and LSTM's sequential dependency handling, the model achieved a 97% accuracy, outperforming traditional methods like ANN and RNN. It also performed well in precision, recall, and F1 score, showcasing its reliability for handling high-dimensional, temporal datasets. While this accuracy is already significant, we aim to enhance it further in future work, refining the model to improve predictive performance. Future improvements include incorporating additional features such as weather data, market conditions, or renewable energy supply trends to enrich the dataset and provide a more comprehensive analysis.

Using metrics like RMSE and MAE, the model proves highly effective for accurate energy market forecasts. Its scalability and adaptability make it suitable for future developments, such as renewable energy integration and policy shifts. This work highlights the advantages of advanced machine learning techniques for electricity price prediction and other time-series or image-driven domains. There may be risk associated with overfitting because of the deep learning models. The overfitting can be reduced by cross- validation or dropout. In the future integrating with smart grids and with renewable energy data like solar and wind power.

**Conflict of Interest**
All the authors do not have any conflict of Interest in this work.

**Data Availability**
All Original research work and study done by all authors and its captured and worked through original resources and no need it involve any third-party materials in this research work along with implementation cum result analysis.